\documentclass[conference]{IEEEtran}
\usepackage{graphicx}
\usepackage{amsmath,amssymb} 
\usepackage{color}
\usepackage{multirow}
\usepackage{multicol}
\usepackage{rotating}
\usepackage[ruled,vlined]{algorithm2e}
\usepackage{cleveref}
\usepackage{adjustbox}

\usepackage{array}
\newcolumntype{P}[1]{>{\centering\arraybackslash}p{#1}}
\newcolumntype{M}[1]{>{\centering\arraybackslash}m{#1}}

\ifCLASSINFOpdf
\else
\fi
\begin{document}
%
\title{DEAL: Deep Evidential Active Learning for Image Classification}

\author{\IEEEauthorblockN{Patrick Hemmer}
\IEEEauthorblockA{Karlsruhe Institute of Technology\\
patrick.hemmer@kit.edu}
\and
\IEEEauthorblockN{Niklas K\"uhl}
\IEEEauthorblockA{Karlsruhe Institute of Technology\\
IBM Germany\\
niklas.kuehl@kit.edu}
\and
\IEEEauthorblockN{Jakob Sch\"offer}
\IEEEauthorblockA{Karlsruhe Institute of Technology\\
jakob.schoeffer@kit.edu}}


%


\maketitle

\begin{abstract}
Convolutional Neural Networks (CNNs) have proven to be state-of-the-art models for supervised computer vision tasks, such as image classification. However, large labeled data sets are generally needed for the training and validation of such models. In many domains, unlabeled data is available but labeling is expensive, for instance when specific expert knowledge is required. Active Learning (AL) is one approach to mitigate the problem of limited labeled data. Through selecting the most informative and representative data instances for labeling, AL can contribute to more efficient learning of the model. Recent AL methods for CNNs propose different solutions for the selection of instances to be labeled. However, they do not perform consistently well and are often computationally expensive. In this paper, we propose a novel AL algorithm that efficiently learns from unlabeled data by capturing high prediction uncertainty. By replacing the softmax standard output of a CNN with the parameters of a Dirichlet density, the model learns to identify data instances that contribute efficiently to improving model performance during training. We demonstrate in several experiments with publicly available data that our method consistently outperforms other state-of-the-art AL approaches. It can be easily implemented and does not require extensive computational resources for training. 
Additionally, we are able to show the benefits of the approach on a real-world medical use case in the field of automated detection of visual signals for pneumonia on chest radiographs.
\end{abstract}


%
\IEEEpeerreviewmaketitle

\section{Introduction}
 
Over the last years, Convolutional Neural Networks (CNNs) have contributed to an unprecedented development of prediction accuracy in the realm of computer vision, even exceeding human-level performance for specific image classification tasks \cite{he2015delving}. However, one major drawback is their dependency on vast amounts of labeled data. Even though more and more data is becoming available, labeling of data instances is often costly. In many application domains, such as medical diagnosis or manufacturing, the knowledge of highly trained experts is essential. This results in the need for developing techniques to reduce data labeling effort, especially when labeling resources are scarce. 
Multiple techniques are available, which lead to a significant decrease in required labeled data, e.g., Inductive Programming \cite{olsson1995inductive}, Semi-Supervised Learning \cite{lin2010semi},  External Memories \cite{vinyals2016matching}, or Active Learning (AL).
The key idea of AL is that a machine learning model can achieve a desired performance level using fewer training instances if it can select the data which is the most beneficial to its learning process \cite{settles2009active}. When applying AL, a model is initially trained on a small labeled data set. Repeatedly, new data instances are selected through an acquisition function, labeled by an expert, and added to the labeled data set until a specific labeling budget is depleted \cite{cohn1996active}.

Intensive research has been conducted on the topic of AL over the past years, and it has been applied successfully to a variety of use cases, such as information extraction from text documents \cite{scheffer2001active} or anomaly detection \cite{ghafoori2019unsupervised}. However, one major challenge in AL is the difficulty to deal with high-dimensional data, such as images \cite{tong2001active}. Multiple researchers have addressed this challenge and developed several methods for the classification of image data with CNNs \cite{beluch2018power,gal2017deep,sener2017active,wang2016cost,yoo2019learning}. 

In the work at hand, we propose Deep Evidential Active Learning (DEAL), an AL algorithm that selects unlabeled data instances for annotation based on prediction uncertainty. Uncertainty estimates are derived by replacing the softmax output function of a CNN with the parameters of a Dirichlet density, as proposed by Sensoy et al. (2018) \cite{sensoy2018evidential}. 
The main contributions of our work are threefold: First, we apply this modified CNN to AL, which enables the model to form a more accurate opinion about the most informative samples. In each AL round, the uncertainty estimates are used to query a new batch of unlabeled data instances for annotation until the labeling resources are depleted. Second, we demonstrate in extensive experiments on MNIST and CIFAR-10 that our proposed method consistently outperforms other state-of-the-art AL approaches. 
Lastly, the application of our method to the use case of detecting visual signals for pneumonia in pediatric chest X-ray images stresses the benefits of DEAL: Its implementation would lead to a 34.52\% reduction in the number of labeled images necessary to achieve a test accuracy of 90\%, in comparison to random acquisition.

Our remaining work is structured as follows: In \Cref{sec:related_work}, we review existing AL approaches for image classification, in particular those compatible with CNNs. Our method's theoretical foundations are outlined and formalized in \Cref{sec:methodology}. In \Cref{sec:experiments}, we conduct a thorough experimental evaluation of DEAL. \Cref{sec:conclusion} concludes our work.

\section{Related Work}\label{sec:related_work}

\begin{table*}[htbp] 
\caption{\label{tab:activelearning} An overview of state-of-the-art AL methods for CNNs.}
\centering
\resizebox{\textwidth}{!}{%
\begin{tabular}{c c | c | c c | c | c | c | c c}
    \hline  
    \multicolumn{2}{c|}{\multirow{4}{*}{Generative}} & \multicolumn{8}{c}{Pool-Based}\\
    \cline{3-10}
    & & \multirow{3}{*}{Diversity} & \multicolumn{2}{c|}{\multirow{3}{*}{Combination}} & \multicolumn{5}{c}{Uncertainty}\\
    \cline{6-10}
    & & & & & \multicolumn{3}{c|}{Non-Ensemble} & \multicolumn{2}{c}{\multirow{2}{*}{Ensemble}}\\
    \cline{6-8}
    & & & & & Softmax & \multicolumn{2}{c|}{Non-Softmax} & & \\
    \hline \hline
    ASAL & cGAN & core-set & \multicolumn{1}{c}{\begin{tabular}[c]{@{}c@{}}Batch\\ BALD\end{tabular}} & BADGE & \multicolumn{1}{c|}{\begin{tabular}[c]{@{}c@{}}minimal\\ margin\end{tabular}} & DEAL & \multicolumn{1}{c|}{\begin{tabular}[c]{@{}c@{}}Learning\\ Loss\end{tabular}} & \multicolumn{1}{c}{\begin{tabular}[c]{@{}c@{}}Deep\\ Ensemble\end{tabular}} & \multicolumn{1}{c}{\begin{tabular}[c]{@{}c@{}}MC-\\ Dropout\end{tabular}}\\
    \hline
    \multicolumn{1}{c}{\begin{tabular}[c]{@{}c@{}}Mayer and Timofte\\ (2020) \cite{mayer2020adversarial}\end{tabular}}  & \multicolumn{1}{c|}{\begin{tabular}[c]{@{}c@{}}Mahapatra et al.\\ (2018) \cite{mahapatra2018efficient}\end{tabular}}   & \multicolumn{1}{c|}{\begin{tabular}[c]{@{}c@{}}Sener and Savarese\\ (2017) \cite{sener2017active}\end{tabular}}   & \multicolumn{1}{c}{\begin{tabular}[c]{@{}c@{}}Kirsch et al.\\ (2019) \cite{kirsch2019batchbald}\end{tabular}}  & \multicolumn{1}{c|}{\begin{tabular}[c]{@{}c@{}}Ash et al.\\ (2019) \cite{ash2019deep}\end{tabular}}  & \multicolumn{1}{c|}{\begin{tabular}[c]{@{}c@{}}Wang et al.\\ (2016) \cite{wang2016cost}\end{tabular}}  & Our work & \multicolumn{1}{c|}{\begin{tabular}[c]{@{}c@{}}Yoo and Kweon\\ (2019) \cite{yoo2019learning} \end{tabular}} & \multicolumn{1}{c}{\begin{tabular}[c]{@{}c@{}}Beluch et al.\\ (2018) \cite{beluch2018power} \end{tabular}} & \multicolumn{1}{c}{\begin{tabular}[c]{@{}c@{}}Gal et al.\\ (2017) \cite{gal2017deep}\end{tabular}} \\
    \hline
\end{tabular}%
}
\end{table*}
AL has been intensively researched over the past decades. Settles (2009) \cite{settles2009active} provides a comprehensive overview of the most commonly used query strategy frameworks. However, it does not take into account approaches compatible with CNNs, as AL research on image data was then predominantly focused on methods such as Support Vector Machines (SVMs). 

In contrast to SVMs, CNNs can capture spatial information in images---the main reason why they have become state-of-the-art technology for image classification tasks \cite{hinton2012improving}, and the primary motivation for researchers to develop AL methods compatible with CNNs. To the best of our knowledge, \Cref{tab:activelearning} summarizes the latest AL approaches for CNNs applied to image data. In the following, we divide them into \textit{generative} and \textit{pool-based} approaches. 

\textit{Generative} methods use Generative Adversarial Networks (GANs) to generate informative samples which are added to the training set. To realize this, Mahapatra et al. (2018) \cite{mahapatra2018efficient} condition their GAN (cGAN) on a real image pool. In contrast, in their method called ASAL, Mayer and Timofte (2020) \cite{mayer2020adversarial} use the generated images to retrieve similar real-world images and add them to the training set after annotation.

\textit{Pool-based} approaches make use of different acquisition strategies to sample the most informative data. We divide them into \textit{diversity-} and \textit{uncertainty-based} approaches as well as a \textit{combination} of both. \textit{Diversity-based} methods pursue the idea of selecting samples that represent the unlabeled data pool most adequately. Sener and Savarese (2017) \cite{sener2017active} frame AL as a core-set selection problem by minimizing the Euclidean distance in the model’s feature space between sampled and non-selected data points. The aim is to query a subset of samples from the data pool such that a model trained on this subset performs comparably to a (hypothetical) model trained on the whole data set. However, distance-based methods like core-set have the disadvantage that distance metrics can concentrate in high-dimensional space, which means that distances between data elements appear identical \cite{franccois2008high}.
The assumption of \textit{uncertainty-based} approaches is that the more uncertain a model is with respect to a prediction, the more informative this data has to be for the model. Wang et al. (2016) \cite{wang2016cost} query the most informative samples by applying least confidence, minimal margin, and entropy acquisition functions to the class probabilities of the softmax output. However, a limitation of this approach is that a model can be uncertain in its predictions even with a high softmax output \cite{gal2016uncertainty}. Therefore, Gal and Ghahramani (2016) \cite{gal2016dropout} address the representation of uncertainty from a Bayesian perspective and propose a framework for modeling uncertainty in deep neural networks with dropout at inference time. It allows to obtain prediction uncertainty estimates by conducting multiple forward passes of each data instance through a model with dropout enabled. Consequently, the models and corresponding prediction results are different for each forward pass. This technique called Monte Carlo (MC-)Dropout yields more accurate uncertainty approximations compared to single softmax point estimates, as it approximates a distribution over the parameters. Applied to AL, it results in the acquisition of more informative samples leading to faster model learning \cite{gal2017deep}. Instead of applying MC-Dropout, Beluch et al. (2018) \cite{beluch2018power} approximate such a distribution through a full model ensemble. Benchmarking against MC-Dropout, they demonstrate that such an ensemble of CNNs can infer more calibrated predictive uncertainties, and thus, further enhances the AL performance. An alternative approach is proposed by Yoo and Kweon (2019) \cite{yoo2019learning}. Their idea is to attach an extra loss prediction model to the network, which learns to predict the losses of unlabeled samples. Consequently, they can query the data points which are expected to have high losses. Moreover, several approaches propose to \textit{combine} diversity- with uncertainty-based data acquisition. As CNNs are trained in a batch setting, selecting instances solely based on uncertainty entails the risk of redundancy in batch-wise queried data, which can lead in some cases to a worse performance than random data selection. Each of the selected points might be informative itself, however, not jointly. In this context, Kirsch et al. (2019) \cite{kirsch2019batchbald} improve the performance of the acquisition function BALD (Bayesian Active Learning by Diverse) \cite{houlsby2011bayesian} in the batch setting (BatchBALD). They query instances by calculating the mutual information between a joint of multiple data points and model parameters. An alternative approach called Batch Active learning by Diverse Gradient Embeddings (BADGE) incorporates both diversity and uncertainty for batch acquisition by measuring uncertainty through gradient embeddings and diversity through sampling instances via the k-MEANS++ initialization scheme \cite{ash2019deep}. 

Our method DEAL belongs to the \textit{uncertainty-based} AL approaches as we acquire each data batch based on uncertainty estimates from a Dirichlet distribution that is placed on the class probabilities. Inferring uncertainty estimates from the softmax probabilities entails the risk that a model can be uncertain in its predictions even with a high softmax output \cite{gal2016uncertainty}, as mentioned earlier. Approaches mitigating this issue require either each data point to be passed multiple times through the network using MC-Dropout \cite{gal2017deep}, infer uncertainty estimates through an ensemble of several models \cite{beluch2018power}, or attach a separate model to the network \cite{yoo2019learning}. The first two options have the drawback that each acquisition step is increasingly time-consuming, whereas the latter involves the implementation overhead of an additional learning loss module.

Acquiring unlabeled data points in each AL round using DEAL has (a) the advantage of deriving high-quality uncertainty estimates leading to faster learning of the model, and (b) requires only one forward pass of each data instance through the network.

\section{Methodology}\label{sec:methodology}

In this section, we first outline the necessary theoretical foundations for our AL approach. Subsequently, we introduce the acquisition function concept and formally describe our method as an algorithm. 

\subsection{Theory of Evidence}

Our AL algorithm is based on the method of quantifying uncertainty in neural networks, as proposed by Sensoy et al. (2018) \cite{sensoy2018evidential}, which originates from the Dempster-Shafer Theory of Evidence (DST) \cite{dempster1968generalization}, a generalization of the Bayesian theory to subjective probabilities. Using subjective logic, DST can be formalized as a Dirichlet distribution, and thus quantify belief masses and uncertainty \cite{jsang2018subjective}. 

In general, a softmax function is typically used in the output layer of CNNs for classification tasks. Specifically, it provides class probability estimates for each class in the form of point estimates. However, a model can be uncertain in its predictions even with a high softmax output for a particular class \cite{gal2016uncertainty}. In contrast to Gal et al. (2017) \cite{gal2017deep} and Beluch et al. (2018) \cite{beluch2018power}, who approximate a distribution of class probabilities through MC-Dropout and model ensembles, we directly model a Dirichlet posterior with its hyperparameters learned from the data. In detail, the idea is to replace the softmax function of the CNN with a nonlinear activation function such as Softsign and use the outputs as evidence vector for a Dirichlet distribution. Moreover, the loss function is adapted in a way that it comprises both the output loss and a regularization term in the form of Kullback-Leibler (KL) divergence, which regularizes the predictive distribution. In the following, we summarize the method to infer prediction uncertainty \cite{sensoy2018evidential}, as this is the basis for our AL algorithm.

First, we can define $K$ mutually exclusive singletons with a non-negative belief mass $b_{ k }$ assignable to each of them, and an overall uncertainty mass $u$. Assuming the singletons to be the $K$ outputs of a CNN ($K$-class classification), we can formulate the following equation with $b_{ k } \geq$  0 for $k = 1,\dots,K$ and $u \geq  0$:
\begin{equation}
u + \sum_{k=1}^{K} b_{k}=1\ .
\end{equation} 
The variable $b_{ k }$ represents the  $k$-th Softsign output and is interpreted as the belief mass of the $k$-th class, whereas $u$ is the uncertainty mass of the particular outputs. Moreover, let $e_{k} \geq 0$ be the evidence for the $k$-th output. Then, the belief mass $b_{ k }$ and uncertainty $u$ with $S = \sum_{i=1}^{K}(e_{i}+1)$ can be defined as:
\begin{equation}\label{eq:equation1}
b_{k} = \frac{e_{k}}{S}\ , \quad  u = \frac{K}{S}\ . 
\end{equation}
In this approach, evidence quantifies the support from data, which results in the classification of a sample into a particular class. Thus, it differs from the Bayesian nomenclature. Additionally, assigning a belief mass corresponds to a Dirichlet distribution with the parameters $\alpha _{k} = e_{k} + 1$. Therefore, a subjective opinion can be formed by the parameters of the corresponding Dirichlet distribution using $b_{k} = (\alpha _{k}-1)/S$. Here, $S = \sum_{i=1}^{K} \alpha _{i}$ represents the Dirichlet strength. Contrary to the standard softmax classifier, which assigns a probability for each possible class and sample, a Dirichlet distribution denotes the density for each probability assignment on the basis of its parameters derived from the evidence vector \cite{jsang2018subjective}. Specifically, a Dirichlet distribution is a probability density function for possible values of the probability mass function $p$. It has $K$ parameters $\alpha = [\alpha_{1},\dots, \alpha _{K}]$ and the form:
\begin{equation}
\begin{aligned}
D(p|\alpha ) &= \begin{cases}
\frac{1}{B(\alpha )}\prod_{i=1}^{K}p_{i}^{\alpha _{i}-1}  & \text{for } p \in S_{K}\ ,\\ 
0 & \text{otherwise },
\end{cases} \\
\textrm{with}\ S_{K} &= \bigg\{
p \, \bigg|\, \sum_{i=1}^{K}p_{i}=1 \text{ and } 0\leq p_{1},\dots, p_{K}\leq 1
\bigg\}\ .
\end{aligned}
\end{equation} 
$B(\alpha )$ is a $K$-dimensional multinomial beta function \cite{korz2000continuous}. In the presence of an opinion, the expected probability for the $k$-th output results from the mean of the respective Dirichlet distribution:
\begin{equation}\label{eq:equation2}
    \widehat{p}_{k} =\frac{ \alpha _{k}}{S}\ .
\end{equation}
A CNN classifies a sample $x_{i}$ with $i \in \{1, \dots, N\}$ by conceiving an opinion as a Dirichlet distribution $D(p_{i}|\alpha_{i} )$, where $p_{i}$ describes the assigned class probabilities. Given a sample $x_{i}$, $f(x_{i}|\Theta)$ is the evidence vector predicted by the CNN with the network parameters $\Theta$. Thus, the parameters of the Dirichlet distribution are $\alpha _{i} = f(x_{i}|\Theta)+1$, and we can calculate the mean $\alpha _{i}/S_{i}$ for estimating the class probabilities. 

For a data sample $x_{i}$, the variable $y_{i}$ denotes the ground-truth class in the form of a one-hot encoded vector. The variable $\alpha _{i}$ represents the parameters of the Dirichlet density on the predictors. Additionally, $D(p_{i}|\alpha _{i})$ is a prior on the likelihood $Mult(y_{i}|\alpha _{i})$, with $Mult(\cdot)$ being a multinominal mass function. Then, the loss function can be defined as follows, using the technique of Type II Maximum Likelihood Estimation:
\begin{equation}
\begin{aligned}
\mathcal{L}_{i}(\Theta) &= - \log\Bigg ( \int \prod_{j=1}^{K}p_{ij}^{y_{ij}} \frac{1}{B(\alpha _{i})}\prod_{j=1}^{K}p_{ij}^{\alpha _{ij}-1}\text{d}p_{i}\Bigg) \\
&= \sum_{j=1}^{K} y_{ij}\Big ( \log(S_{i})-\log(\alpha _{ij}) \Big )\ .
\end{aligned}
\end{equation}
To ensure that the total evidence decreases to zero for a sample that cannot be correctly classified, the KL-divergence $KL(\cdot)$ is incorporated into the loss function with $\lambda _{t} = \min(1,t/10) \in \left [ 0,1 \right ]$, where $t$ denotes the current training epoch:  
\begin{equation}
\mathcal{L}(\Theta )=\sum_{i=1}^{N}\mathcal{L}_{i}(\Theta) + \lambda_{t}\sum_{i=1}^{N}KL\Big( D(p_{i}|\widetilde{\alpha}_{i})\, \Big|\Big|\, D(p_{i}| 1)  \Big)\ .
\end{equation}
The term $D(p_{i}|1)$ refers to the uniform Dirichlet distribution, and $\widetilde{\alpha}_{i} = y_{i} + (1-y_{i}) \odot \alpha_{i}$, with $\odot$ referring to the Hadamard (element-wise) product. We can calculate the KL-divergence as follows, with $\Gamma(\cdot)$ denoting the gamma function and $\psi(\cdot)$ referring to the digamma function:
\begin{equation}
\begin{aligned}
KL\Big( D(p_{i}|\widetilde{\alpha}_{i})\, \Big|\Big|\, D(p_{i}|1) \Big)
= \log \Bigg( \frac{\Gamma \Big(\sum_{k=1}^{K} \widetilde{\alpha}_{ik}\Big)}{\Gamma\big(K\big)\prod_{k=1}^{K}\Gamma \big(\widetilde{\alpha}_{ik}\big)} &\Bigg) \\ 
+ \sum_{k=1}^{K}\Big(\widetilde{\alpha}_{ik} - 1\Big) \Bigg( \psi \Big(\widetilde{\alpha}_{ik}\Big) - \psi  \bigg( \sum_{j=1}^{K}\widetilde{\alpha}_{ij} \bigg) &\Bigg)\ .
\end{aligned}
\end{equation}
A CNN with these modifications is the basis for our proposed AL algorithm. 


\subsection{Uncertainty-Based AL}\label{sec:uncertainty-based-AL}

Before formally defining our AL algorithm, we introduce the concept of an acquisition function including the minimal margin uncertainty measure. 

\subsubsection{Acquisition Function} Given a model $M$, an unlabeled data pool $D_{l=0}$, a labeled data pool $D_{l=1}$ and observations $x\in D_{l=0} $, an AL algorithm uses an acquisition function $a(x,M)$ to choose the next data sample(s) to be queried \cite{gal2017deep}. We define $x^{\star}$ as the immediate next sample to be queried, satisfying
\begin{equation}\label{eq:formula_acquisition_function}
    x^{\star} = \text{argmax}_{x\in D_{l=0}}\, a(x,M)\ .
\end{equation}
For $a(x,M)$, we apply the uncertainty measure \textit{minimal margin}: Choose the sample with the smallest
\begin{equation}\label{eq:minimal_margin}
\begin{aligned}
margin(y|x,D_{l=1}) =  p(y = k_{1}|x,D_{l=1}&)\\
-p(y = k_{2}|x,D_{l=1}&)\ ,
\end{aligned}
\end{equation}
where $k_{1}$ and $k_{2}$ are the first and second most probable class labels of the respective sample.

The CNN's expected class probabilities serve as input, derived from \Cref{eq:equation2}. Besides the uncertainty measure minimal margin, other metrics such as Shannon Entropy \cite{shannon1948mathematical} are applicable as well. However, we have found in previous experiments that the DEAL algorithm yields the best results with the minimal margin measure.

\subsubsection{AL Framework} The pool-based AL setting originates from an unlabeled data pool $D_{l=0}$. Initially, a model $M$ is trained on a small data set $D_{l=1}$, which is drawn uniformly at random. Labels are obtained from an expert. In each AL round, $A$ data samples are selected for labeling and added to $D_{l=1}$, based on the acquisition function $a(x,M)$. Subsequently, the model is trained from scratch. This process is repeated until a given labeling budget is exhausted.

Deriving classification uncertainty by placing a Dirichlet distribution on the class probabilities in combination with the minimal margin uncertainty measure forms the basis of DEAL. We formalize this approach in \Cref{DEAL_algorithm}.
\begin{center}
\begin{algorithm}
\SetAlgoLined
\textbf{Input:} Unlabeled set $D_{l=0}$, labeled set $D_{l=1}=\emptyset$, model $M$ with loss function $\mathcal{L}$, acquisition size $A$, labeling budget $B$ (\#samples).

\KwResult{Updated model $M_{N}(\Theta_{N})$.}
Acquisition function based on \Cref{eq:minimal_margin}\; 
Compute $N = \min(\left \lfloor B/A \right \rfloor, \left \lfloor |D_{l=0}|  \right /A \rfloor)$\;
Set $D_{l=1} \leftarrow A$ samples drawn uniformly at random from $D_{l=0}$, labeled by expert\;
Set $D_{l=0} \leftarrow D_{l=0}\setminus D_{l=1}$\;
Initialize parameters $\Theta_{0}$ and train $M_{0}(\Theta_{0})$ on $D_{l=1}$
minimizing $\mathcal{L}$\;
\For{$t = 1, 2,\dots, N$}{
\For{$i = 1, 2,\dots, A$}{
    Compute $x^{\star}$ as in \Cref{eq:formula_acquisition_function}\;
    Request ground-truth label for $x^{\star}$\;
    Set $D_{l=1} \leftarrow D_{l=1}\cup \{x^{\star}\}$\; 
    Set $D_{l=0} \leftarrow D_{l=0} \setminus \{x^{\star}\}$\;
 }


Initialize $\Theta_{t}$ and train $M_{t}(\Theta_{t})$ on $D_{l=1}$
minimizing $\mathcal{L}$\;
 }
 \caption{Deep Evidential Active Learning (DEAL)}
 \label{DEAL_algorithm}
\end{algorithm}
\end{center}

\section{Experiments}\label{sec:experiments}

In this section, we  first present the experimental scenario used for the evaluation of DEAL on MNIST \cite{lecun1998gradient} and CIFAR-10 \cite{krizhevsky2009learning}. We benchmark the performance of DEAL against other state-of-the-art AL approaches and highlight its advantage in terms of acquisition time. Second, we present the results of DEAL applied to a real-world medical use case in the field of automated detection of visual signs for pneumonia on chest radiographs.

\begin{figure*}[!htbp]
\begin{multicols}{2}
    \centering
    \includegraphics[width=0.64\linewidth]{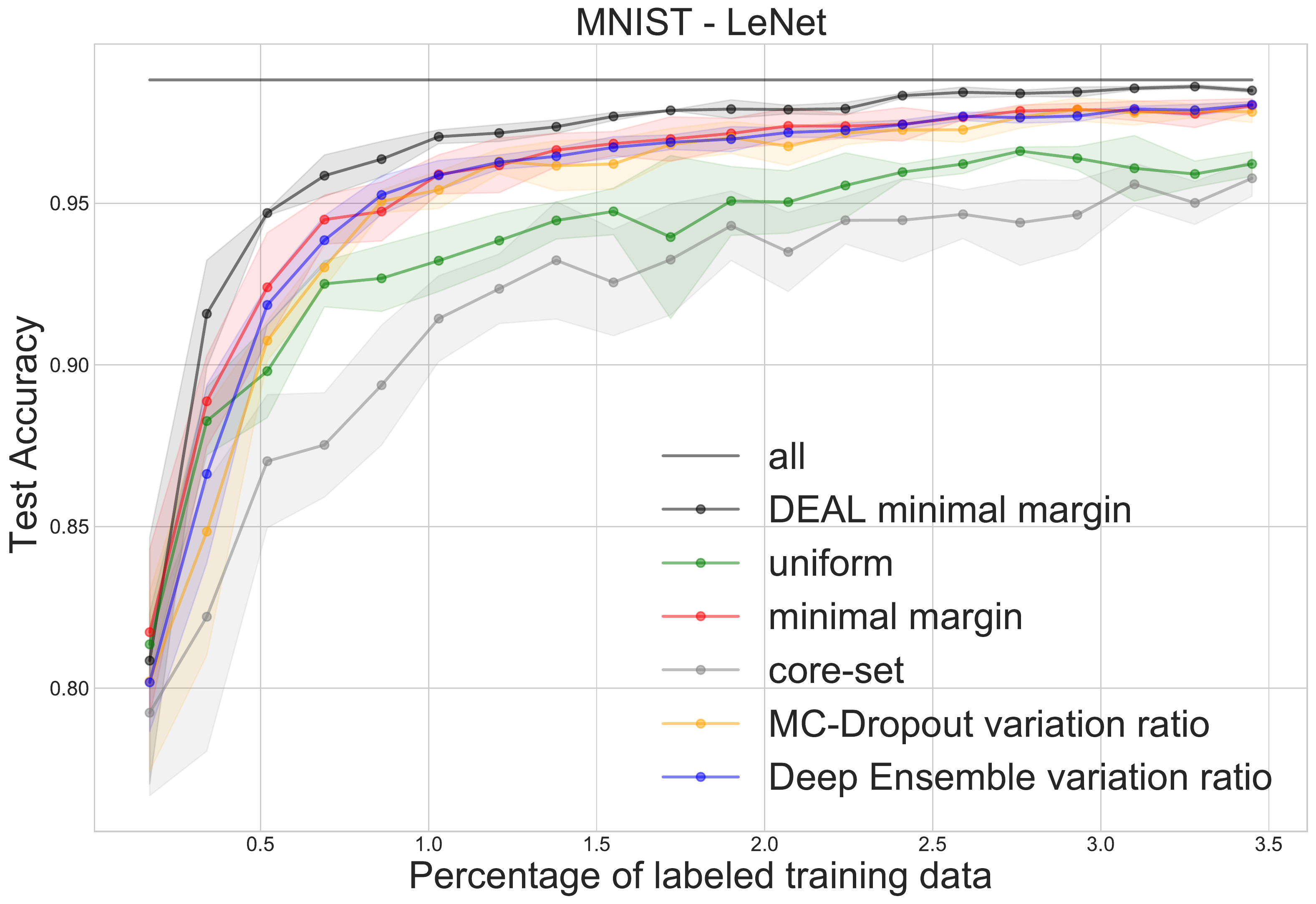}\par 
    \includegraphics[width=0.64\linewidth]{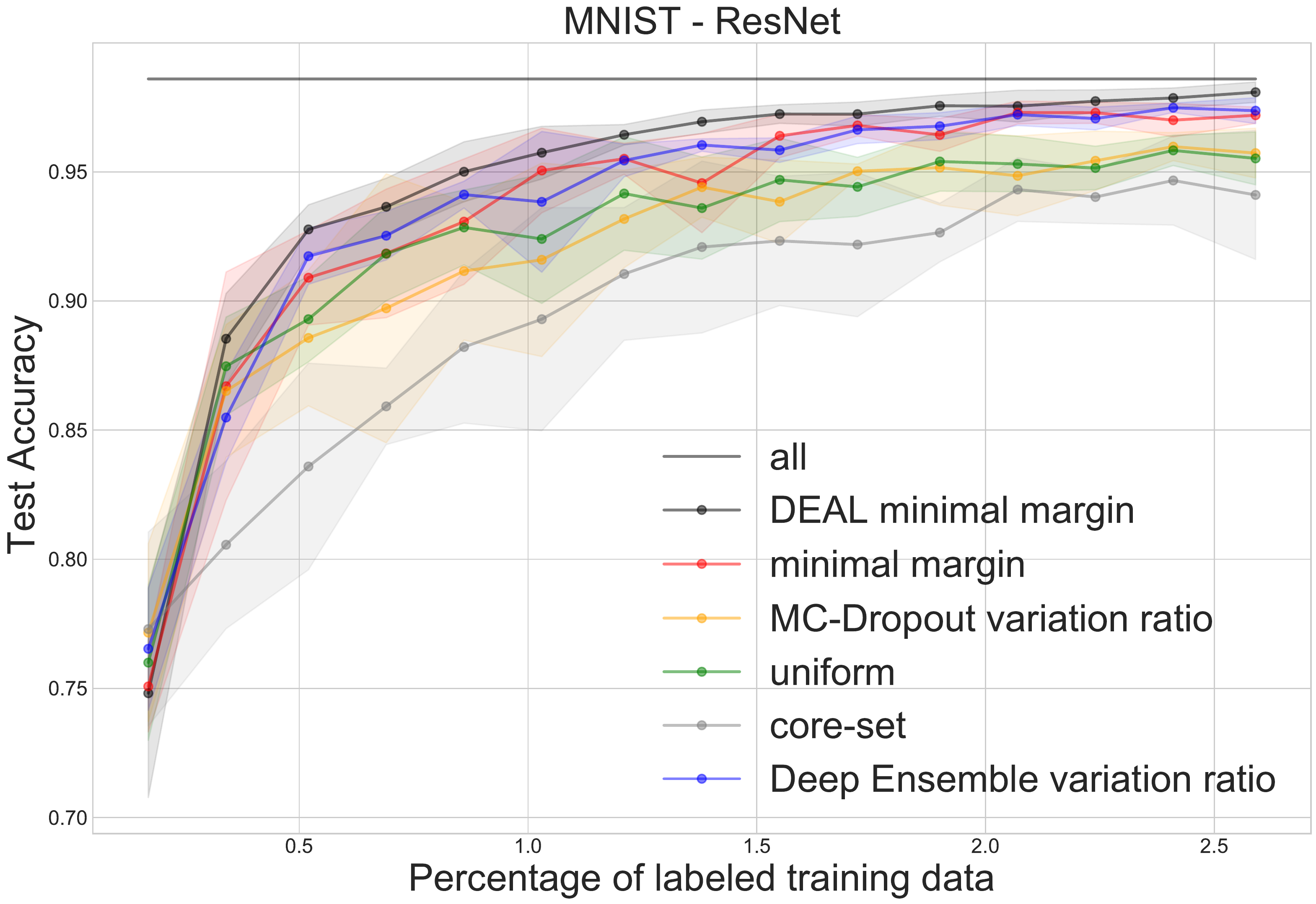}\par 
    \end{multicols}
\begin{multicols}{2}
    \centering
    \includegraphics[width=0.64\linewidth]{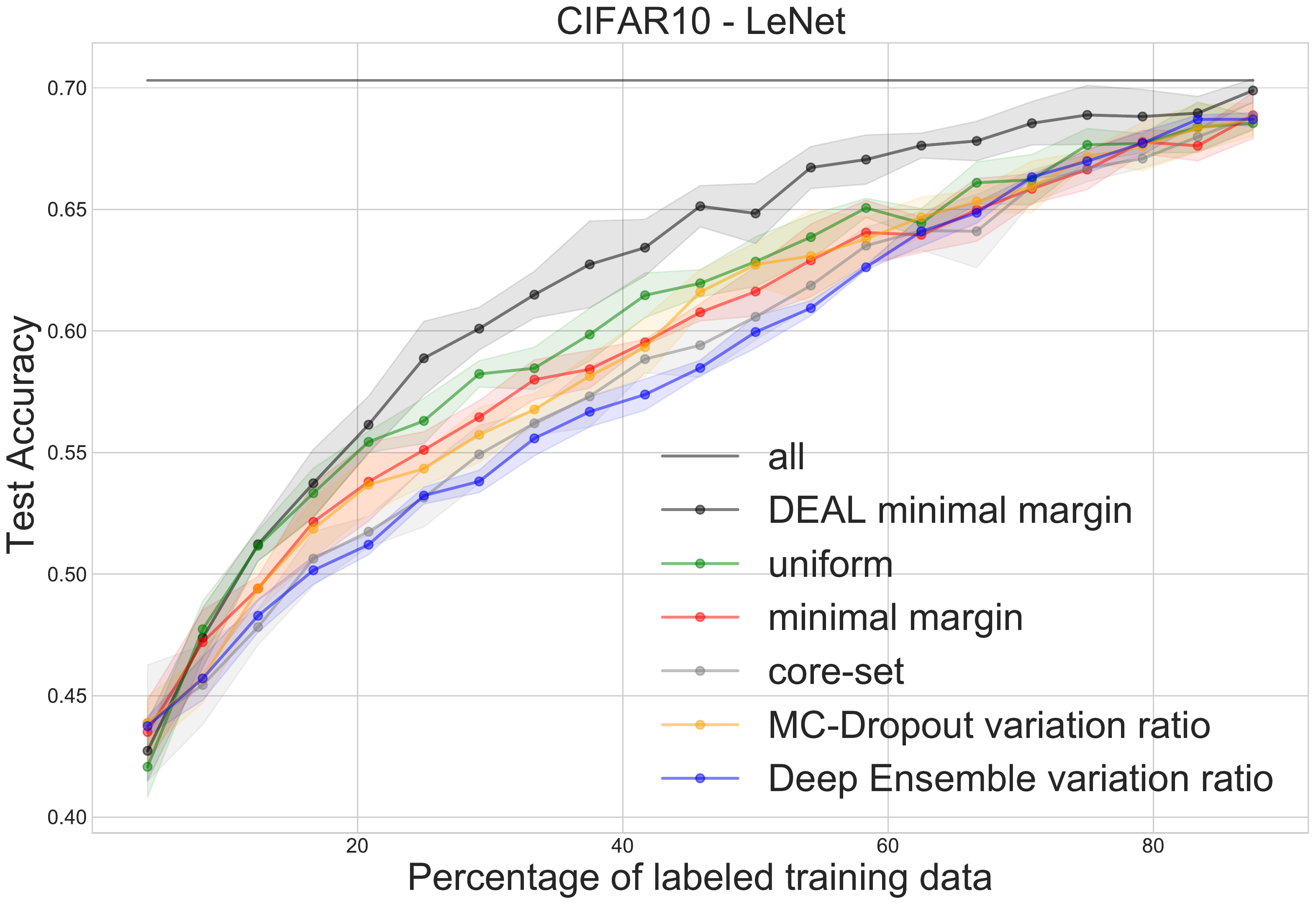}\par
    \includegraphics[width=0.64\linewidth]{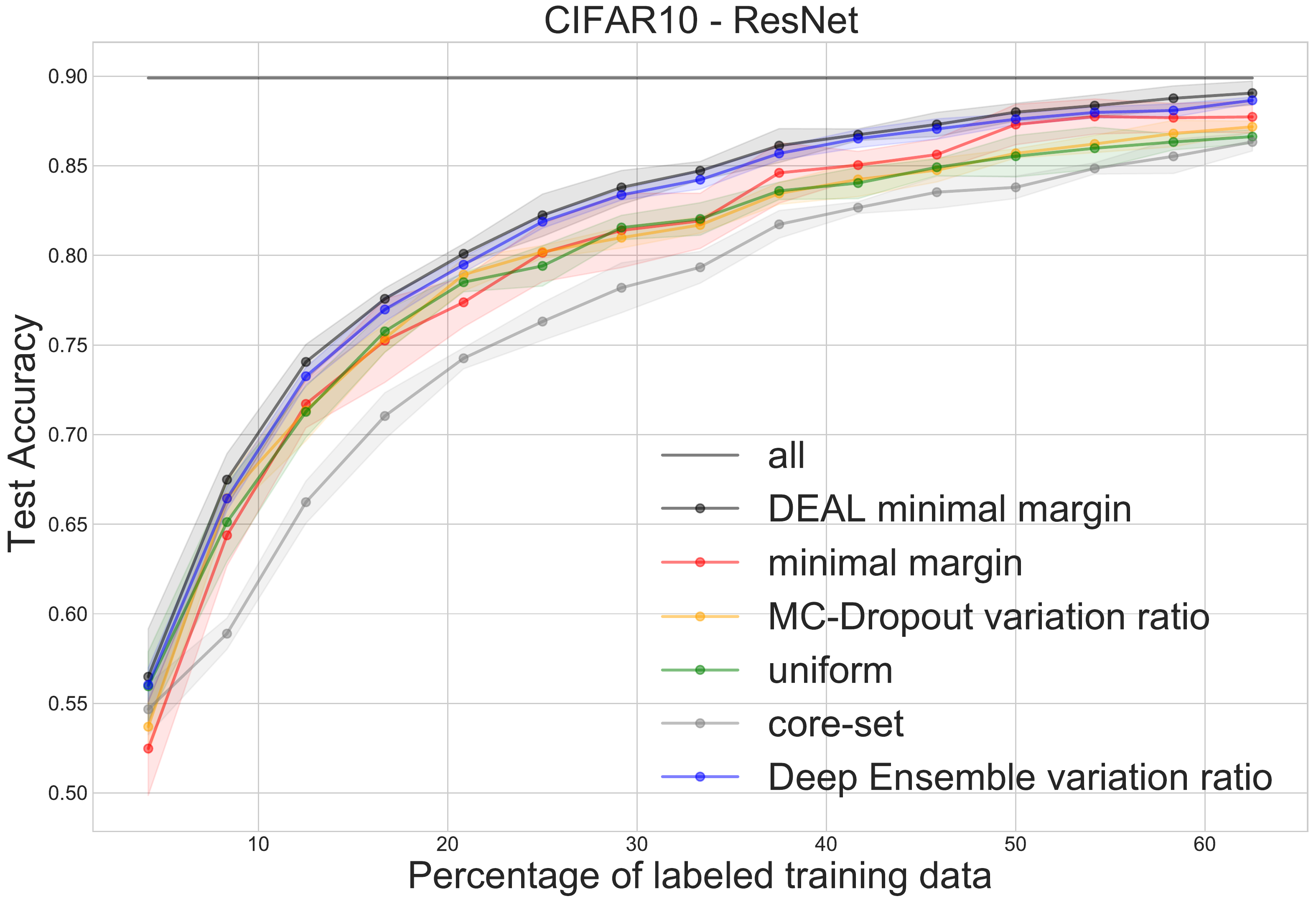}\par
\end{multicols}
\caption{MNIST (upper two) and CIFAR-10 (lower two) test accuracy over the percentage of acquired labeled training data. We benchmark the performance of DEAL against the approaches introduced in \Cref{sec:implementation_details}. The solid horizontal line represents the CNN trained with all labeled training data. Shaded regions display standard deviations.}
\label{fig:edl_MNIST_CIFAR10_comparison_benchmarks}
\end{figure*}

\subsection{Implementation Details}\label{sec:implementation_details}

We evaluate our method on the publicly available scientific data sets MNIST and CIFAR-10. The former comprises a total of 70,000 greyscale images, of which we assign 58,000 to the training set, 2,000 to the validation set, and 10,000 to the test set. The latter consists of 60,000 RGB-images with 48,000 belonging to the training set, 2,000 to the validation set, and 10,000 to the test set. Both data sets involve ten classes each. We choose these two data sets for evaluation of our approach as they differ in terms of image diversity: While MNIST contains inherently many redundant images, CIFAR-10's images are more diverse \cite{vodrahalli2018all}. 

We conduct our experiments with two popular CNN architectures on both data sets. The first one is the LeNet \cite{lecun1999object}
and the second one the ResNet-18 \cite{he2016deep} implementation.
For both architectures, the standard softmax layer is replaced by a Softsign layer whose output is used as evidence vector for the Dirichlet distribution. 
In each AL round, we train the networks from scratch for 100 epochs, using batch size 32 for the LeNet, 8 for the ResNet on MNIST and 64 on CIFAR-10. We choose a learning rate of 0.0005 and Adam \cite{kingma2014adam} as optimizer for both networks. All experiments are implemented in TensorFlow \cite{abadi2016tensorflow}. Our code is available at \texttt{github.com/DeepLearningResearch/DEAL}. The experiments are conducted in the following standardized setting: In the first AL round, we sample uniformly at random 100 MNIST and 2,000 CIFAR-10 images and train the model over the defined number of epochs. In each subsequent AL round, a new batch of images with the same acquisition size is selected, added to the labeled data pool, and the model is re-trained from scratch. 
We repeat this procedure until the test accuracy differs only marginally from that of a model trained on all labeled images. Regarding the LeNet architecture on MNIST, we stop the AL algorithm after the acquisition of 2,000 and on CIFAR-10 after training with 42,000 images. With the ResNet architecture we terminate the AL algorithm after the selection of 1,500 MNIST and 30,000 CIFAR-10 images.
Each experiment is repeated 5 times, and the average test accuracy including standard deviation is reported.

We compare our AL method with the following state-of-the-art approaches: Minimal margin with softmax \cite{wang2016cost}, core-set \cite{sener2017active}, MC-Dropout \cite{gal2017deep} and Deep Ensemble \cite{beluch2018power}. The latter two use both the variation ratio \cite{freeman1965elementary} as acquisition function. Both Gal et al. (2017) \cite{gal2017deep} and Beluch et al. (2018) \cite{beluch2018power} apply further uncertainty-based acquisition functions to their approach. However, we only use variation ratio as it achieves the best AL performance in both papers. Random sampling serves as a baseline for all. Analogously to Beluch et al. (2018) \cite{beluch2018power}, for MC-Dropout, we conduct 25 forward passes, and for Deep Ensemble, each ensemble consists of 5 models with different random initializations.

\begin{table*}[!htbp]
\caption{T-statistics of DEAL paired with other AL methods considering the respective model architectures and data sets. The asterisk denotes statistical significance at the 0.01 level.}
\label{tab:t-statistic}
\resizebox{\textwidth}{!}{%
\begin{tabular}{M{2.5cm}M{2.3cm}M{2.3cm}|M{2.3cm}M{2.3cm}M{2.3cm}M{2.3cm}M{2.3cm}}
\hline
T-statistic & Model architecture & Data set & uniform & minimal margin & core-set & MC-Dropout variation ratio & Deep Ensemble variation ratio \\ \hline \hline
\multirow{4}{*}{\begin{tabular}[c]{@{}c@{}}DEAL \\ minimal \\ margin\end{tabular}} & \multirow{2}{*}{LeNet} & MNIST & 11.5612$^{*}$  & 5.6201$^{*}$  &   10.7127$^{*}$   & 4.5650$^{*}$  & 5.1262$^{*}$ \\
 &  & CIFAR-10 & 7.1938$^{*}$ & 8.2872$^{*}$ & 8.6366$^{*}$ & 7.9109$^{*}$ & 7.4767$^{*}$  \\
 & \multirow{2}{*}{ResNet} & MNIST & 7.5776$^{*}$ & 5.5596$^{*}$ & 7.0798$^{*}$ & 6.3425$^{*}$ & 3.4235$^{*}$ \\
 &  & \multicolumn{1}{c|}{CIFAR-10} & \multicolumn{1}{c}{15.0278$^{*}$} & \multicolumn{1}{c}{8.4264$^{*}$} & \multicolumn{1}{c}{10.1137$^{*}$} & \multicolumn{1}{c}{15.0638$^{*}$} & \multicolumn{1}{c}{9.0240$^{*}$} \\ \hline
\end{tabular}%
}
\end{table*}

\begin{table*}[htbp]
\caption{Average number of images over 5 experimental runs to achieve a predefined model performance (ResNet) on the test set. The values in parentheses denote standard deviations.}
\label{tab:number_images_test_performance}
\resizebox{\textwidth}{!}{%
\begin{tabular}{M{2.5cm}|M{2.3cm}M{2.3cm}M{2.3cm}M{2.3cm}M{2.3cm}M{2.3cm}}
\hline
Test accuracy & uniform & minimal margin & core-set & MC-Dropout variation ratio & Deep Ensemble variation ratio & DEAL minimal margin \\ \hline \hline
95\% (MNIST) & 820 (160) & 620 (75) & 1,280 (160) & 860 (136) & 680 (75) & 540 (49) \\

87\% (CIFAR-10) & 28,000 (2,000) & 22,500 (2,958) & 30,000 (0) & 29,000 (1,000) & 22,800 (980) & 21,200 (2,040) \\
\hline
\end{tabular}%
}
\end{table*}

\begin{table*}[!htbp]
\caption{Average acquisition time in seconds over all AL rounds for MNIST and CIFAR-10 (ResNet). The experiments are conducted on 
a Tesla V100-SXM2-32GB. The number of forward passes for MC-Dropout is denoted by $n_{fp}$, whereas $n_{en}$ refers to the number of ensemble members. Moreover, $t_{en}$ is the average training time for one member over all epochs. We use the same setting as specified in \Cref{sec:implementation_details}.}
\label{tab:runtime_analysis}
\resizebox{\textwidth}{!}{%
\begin{tabular}{M{2cm}|M{2cm}M{2.3cm}M{2cm}M{2.5cm}M{2.8cm}M{2.2cm}}
\hline
Data set & uniform & minimal margin & core-set & MC-Dropout variation ratio & Deep Ensemble variation ratio & DEAL minimal margin \\ \hline \hline
MNIST & 1.38 & 12.62 & 23.41 & 1.90$n_{fp}+$18.37 & $t_{en} n_{en} +$18.37 & 12.54 \\
CIFAR-10 & 17.22 & 139.25 & 1070.79 & 1.90$n_{fp} +$143.33 & $t_{en} n_{en} +$143.33 & 139.51 \\ \hline
\end{tabular}%
}
\end{table*}

\subsection{Experimental Results}\label{sec:experimental_results}
We benchmark DEAL with minimal margin-based acquisition function against the state-of-the-art approaches from \Cref{sec:implementation_details} using both networks. 
The test accuracy for MNIST is illustrated in the upper two graphs of \Cref{fig:edl_MNIST_CIFAR10_comparison_benchmarks}, whereas the lower two display the results for CIFAR-10. 
For both networks and data sets, DEAL consistently outperforms all other approaches over all acquisition rounds. In detail, averaged over all 5 experiments and all AL rounds, on MNIST, DEAL outperforms the second-best method by 1.01\% (LeNet: minimal margin with softmax) and 1.06\% (ResNet: Deep Ensemble), respectively. Concerning CIFAR-10 using LeNet, none of the other approaches yields better test accuracy than random sampling. However, DEAL outperforms this baseline, on average by 1.51\%. Regarding the ResNet network, the Deep Ensemble approach yields the second best test accuracy. Here, DEAL achieves an average improvement of 0.51\% relative to this method.

In order to demonstrate the statistical significance of these findings, we perform a paired t-test, where the pairs consist of test accuracy development over all AL rounds for DEAL and each benchmark method. \Cref{tab:t-statistic} displays all t-statistics. It becomes evident that for both network architectures and data sets the t-statistic indicates statistical significance at the 0.01 level.

In practice, it is often essential to reach a predefined performance threshold with as little labeling effort as possible. Thus, we illustrate in \Cref{tab:number_images_test_performance} the number of images that are required to achieve a predefined test set accuracy. It highlights the advantages of AL in general and DEAL in particular. For instance, with the help of our approach, a labeling expert would have to label 280 images less for the MNIST and 6,800 images less for the CIFAR-10 data set (compared to random sampling), in order to achieve a test set accuracy of 95\% and 87\%, respectively. This means that the annotation effort can be reduced by 34.15\% for MNIST and 24.29\% for CIFAR-10. Lastly, the comparison with the second-best approach at this selected accuracy level---minimal margin with softmax---shows that for MNIST, 80 images less have to be labeled, which corresponds to a saving of 12.91\%. In contrast, CIFAR-10 requires 1,300 fewer images, which is a reduction of 5.78\%.

\subsubsection{Acquisition Time Analysis}

Even though computing capacities are ever-increasing, the objective of an AL algorithm should not be limited to reaching a desired performance level with fewer training data, but also to make the selection of the next batch of data samples as time-efficient as possible. Thus, we compare the acquisition time of DEAL with that of the other methods. For each approach, we calculate the average sampling time in seconds (s) over all acquisition rounds for the MNIST and CIFAR-10 AL settings, respectively.
The time measurements are displayed in \Cref{tab:runtime_analysis}. With 12.54s for MNIST and 139.51s for CIFAR-10, the acquisition step within DEAL is one of the least time-consuming. Only random sampling with 1.38s (MNIST) and 17.22s (CIFAR-10) requires less time. Minimal margin with 12.62s (MNIST) and 139.25s (CIFAR-10) takes approximately the same amount of acquisition time.
In contrast, the run-time of the acquisition step for MC-Dropout depends on the number of forward passes $n_{fp}$ through the network at test time. Using $n_{fp}=$ 25, it results in 65.87s (MNIST) and 190.83s (CIFAR-10), respectively. Similarly, without parallel training of the ensemble members for Deep Ensemble, the acquisition step depends in time on the number of members $n_{en}$ and the average training time $t_{en}$ over all epochs per member. Therefore, it is dependent on the network and data dimensionality. With $n_{en}=$ 5, $t_{en}=$ 718.69s for MNIST and $t_{en}=$ 2,253.39s for CIFAR-10, the average MNIST acquisition time is 3,611.82s, whereas for CIFAR-10, it results in 11,410.28s. Lastly, the core-set approach is dependent on the dimensionality of the data instances to be queried as well. While the acquisition run-time on MNIST is, on average, only 10.87s slower in comparison with DEAL, the difference of 931.28s becomes noticeable on CIFAR-10. 

\begin{table*}[!htbp]
\caption{T-statistics of DEAL paired with other AL methods on the pediatric pneumonia data set. The asterisk denotes statistical significance at the 0.01 level.}
\label{tab:t-statistic_pneumonia}
\resizebox{\textwidth}{!}{%
\begin{tabular}{M{2.5cm}|M{2.3cm}M{2.3cm}M{2.3cm}M{2.3cm}M{2.3cm}}
\hline
T-statistic &  uniform & minimal margin & core-set & MC-Dropout variation ratio & Deep Ensemble variation ratio \\ \hline \hline
DEAL minimal margin & 7.3670$^{*}$  & 6.6137$^{*}$  &   9.8789$^{*}$   & 9.1429$^{*}$  & 9.2657$^{*}$ \\ \hline
\end{tabular}%
}
\end{table*}

\subsubsection{Real-World Data Set}
We assess the proposed approach on a real-world medical use case in the field of automated pneumonia detection from chest X-rays. Pneumonia is an infection of the lung, which involves a higher global mortality rate among young children than any other infectious disease \cite{rudan2008epidemiology}. In fact, pneumonia is causing death to more children than HIV/AIDS, malaria, and measles combined \cite{adegbola2012childhood}. In the United Stated, about 50,000 people die from pneumonia each year \cite{pneumonia}. Chest radiographs are the most commonly used method for the diagnosis of this disease. However, its interpretation requires knowledge and experience of highly trained radiologists. Additionally, immediate radiologic interpretation of images is not always available, especially in regions with deficient medical infrastructure. Thus, an automated diagnosis system could not only support radiologists in image interpretation but also transfer knowledge to regions with missing expertise. Rajpurkar et al. (2017) \cite{rajpurkar2017chexnet} and Varshni et al. (2019) \cite{varshni2019pneumonia} successfully demonstrate the applicability of CNNs to pneumonia detection from chest X-rays. However, one general challenge for building such systems in the medical domain remains the availability of annotated data instances. In order to demonstrate the benefits of our proposed approach for reducing the necessary number of annotated images, we apply DEAL to a pediatric pneumonia data set collected by Kermany et al. (2018) \cite{kermany2018identifying}. From the original data set with 5,232 X-ray images of children, we randomly sample a subset of 3,100 images consisting of equal partitions of images with clear lungs and images with visual symptoms for pneumonia. We do this to achieve better comparability with the test accuracy development of the evenly balanced scientific data set analyses. 
\begin{figure}[htbp]
  \centering
  \begin{multicols}{2}
    \centering
    \includegraphics[width=\linewidth]{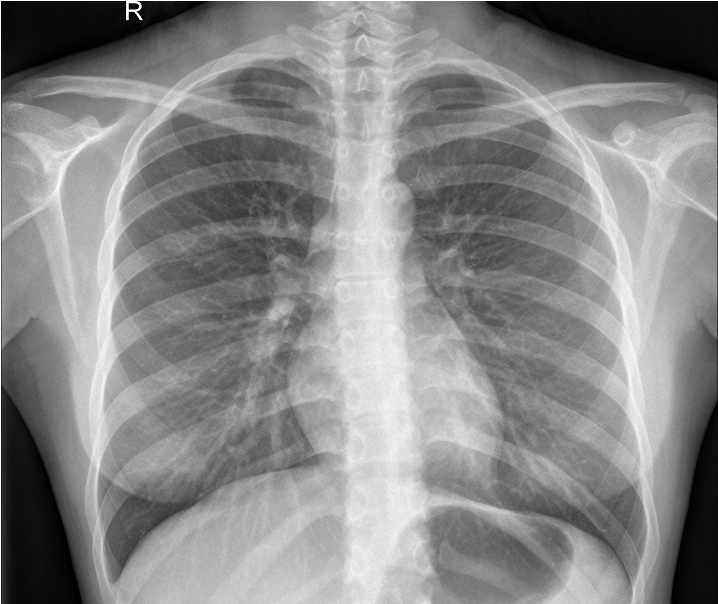}\par
    \includegraphics[width=\linewidth]{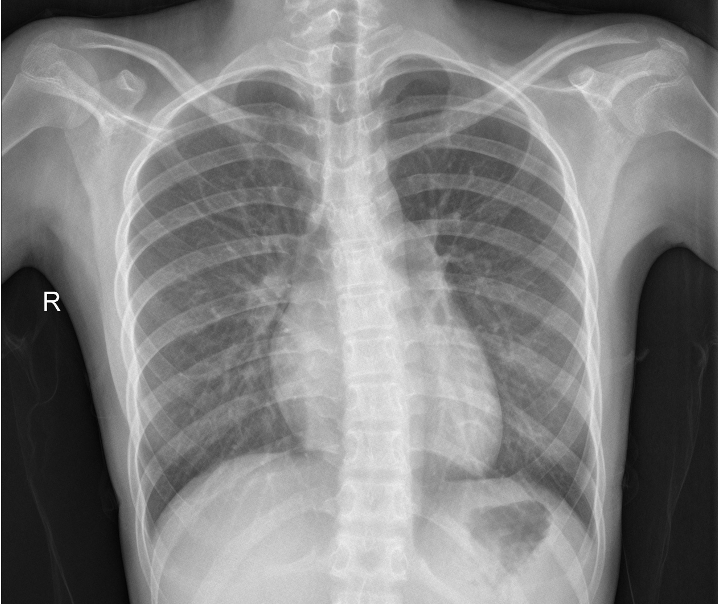}\par 
    \end{multicols}
\begin{multicols}{2}
    \centering
    \includegraphics[width=\linewidth]{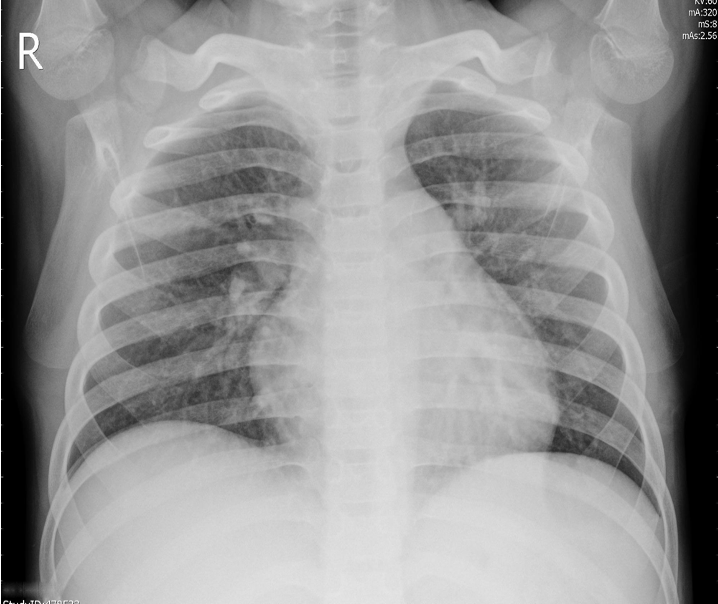}\par
    \includegraphics[width=\linewidth]{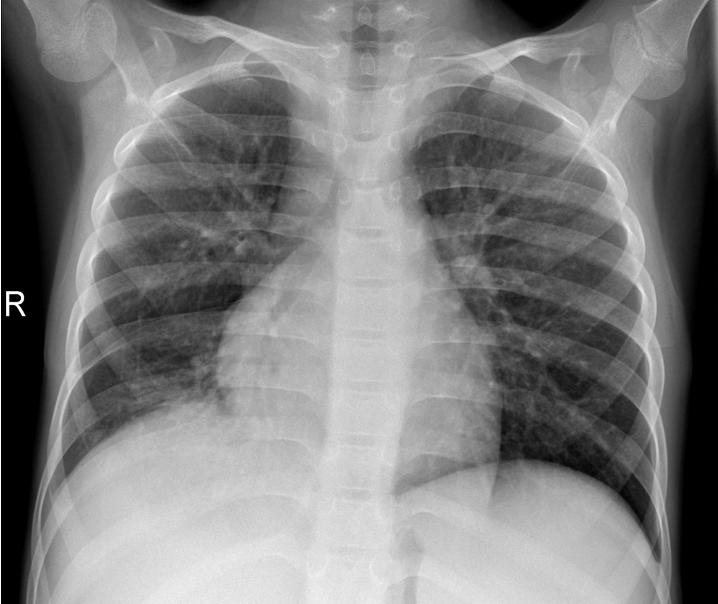}\par
\end{multicols}
  \caption{The upper two chest X-rays represent lungs without any abnormal opacification in the image. The lower left chest X-ray depicts a viral pneumonia, whereas the lower right image shows a chest X-ray of a bacterial pneumonia.}
  \label{fig:exemplary_pneumonia}
\end{figure}

\Cref{fig:exemplary_pneumonia} exemplifies two healthy lungs (upper two) and two lungs suffering from pneumonia (lower two). All images of the data set are recorded with different resolutions. Therefore, we convert the images to greyscale and compress them to 128$\times$128 pixels. We allocate 1,500 images to the training set, 200 to the validation set, and 1,400 to the test set. 
\begin{figure}[htbp]
 \centering
\includegraphics[width=0.8\linewidth]{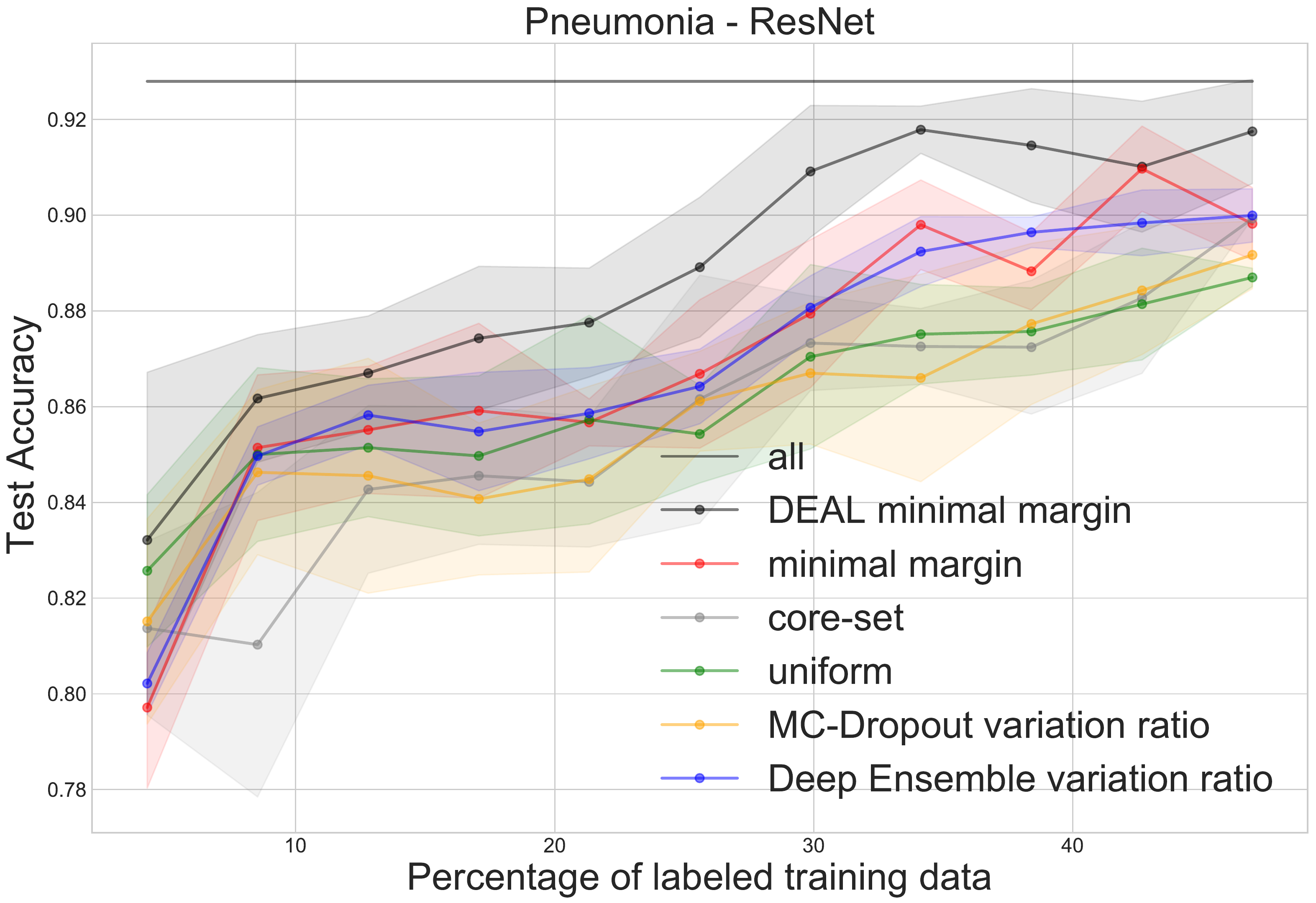}\par
\caption{Pediatric pneumonia data set test accuracy over the percentage of acquired labeled training data. We benchmark the performance of DEAL against the approaches introduced in \Cref{sec:implementation_details}. The solid horizontal line represents the CNN trained with all labeled training data. Shaded regions display standard deviations.}
\label{fig:figure_real_world_use_case}
\end{figure}

We conduct all experiments with the ResNet architecture introduced in \Cref{sec:implementation_details}. In each AL round, the network is trained from scratch for 100 epochs with a batch size of 8. We use a learning rate of 0.0005 and the Adam optimizer. Initially, we train the model with 64 randomly sampled images. In each AL round, 64 images are queried according to the respective acquisition function. We terminate the AL algorithm after the selection of 704 images, as the test accuracy of a model trained on all images is only slightly superior.

\Cref{fig:figure_real_world_use_case} displays the experimental results of DEAL in comparison with all benchmarks. Similar to the analysis on MNIST and CIFAR-10, we note that DEAL consistently outperforms all other approaches. 

More precisely, averaged over all 5 experiments and all AL rounds, DEAL outperforms the second-best method---minimal margin with softmax---by 1.76\%, and the baseline random sampling by 2.86\%. To achieve a test accuracy of 90\%, an expert using DEAL could reduce the number of images to be labeled by 64 compared to the second-best approach, and by 243 images in comparison to the baseline random sampling. This equals a saving of 12.19\% and 34.52\%, respectively. When performing the paired t-test (\Cref{tab:t-statistic_pneumonia}), again, we find that all tested pairs---DEAL and each benchmark method---indicate statistical significance at the 0.01 level. Finally, it is noteworthy that all methods have a larger standard deviation over the individual runs compared to the experimental results on MNIST and CIFAR-10. We suspect that the reason for this might be the small number of images used to train the CNN. 
 
\section{Conclusion}\label{sec:conclusion}

In this paper, we present a novel uncertainty-based AL method and demonstrate its ability to outperform state-of-the-art AL approaches on the scientific data sets MNIST and CIFAR-10. By inferring uncertainty estimates from unlabeled data instances obtained from the class probabilities of a Dirichlet distribution instead of the standard softmax output, data points that contribute to more efficient learning can be identified. We show that to achieve a predefined model performance, our approach not only reduces the number of required labeled data instances but is also superior in terms of acquisition time. By applying our method to a real-world use case of pediatric pneumonia chest X-ray images, we point out its potential to successfully reduce the number of images that have to be labeled by experts. 
A shortcoming of this approach may be found in the solely uncertainty-based selection of data points, since the acquisition exclusively on the basis of uncertainty carries the risk of introducing redundancy in each acquired batch, which can lead to suboptimal solutions. Hence, an interesting avenue for future research would be to incorporate a diversity criterion into our approach.

\bibliographystyle{splncs04}
\bibliography{bibliography}


%

\end{document}